\makeatletter
\def\thanks#1{\protected@xdef\@thanks{\@thanks
        \protect\footnotetext{#1}}}
\makeatother

\documentclass{article}

 \PassOptionsToPackage{numbers, compress}{natbib}
\usepackage[final]{nips_2018}
 \usepackage{booktabs,lipsum,xfrac} 
 \usepackage{graphicx}
 \usepackage{multirow}
\usepackage{rotating}
\usepackage{dblfloatfix}
\usepackage{bbm}
 \usepackage{array}
\graphicspath{{img/}}
\usepackage[utf8]{inputenc} % allow utf-8 input
\usepackage[T1]{fontenc}    % use 8-bit T1 fonts
\usepackage{hyperref}       % hyperlinks
\usepackage{url}            % simple URL typesetting
\usepackage{booktabs}       % professional-quality tables
\usepackage{amsfonts}       % blackboard math symbols
\usepackage{nicefrac}       % compact symbols for 1/2, etc.
\usepackage{microtype}      % microtypography
\usepackage{tabularx}

\usepackage{wrapfig,blindtext}
\usepackage[none]{hyphenat}
\usepackage{comment}
\usepackage{enumitem}
\setlist{itemsep=0pt, topsep=0pt}

\newlist{myenumi}{description}{10}
\setlist[myenumi]{labelindent=\parindent, leftmargin=*, label=(\roman*), align=left}
\setlist[myenumi]{leftmargin=0pt}

\usepackage{tikz}
\usetikzlibrary{shapes,shapes.multipart}

\usepackage{amsmath}
\usepackage[nodisplayskipstretch]{setspace}
\usepackage[justification=centering]{caption}

\usepackage{multicol}

\usepackage{amsmath, amsfonts, siunitx}
\usepackage{amssymb}
\usepackage{balance}
\usepackage{lscape}
\usepackage{adjustbox}

\usepackage{titlesec}

\titlespacing\section{0pt}{0pt plus 0pt minus 0pt}{0pt plus 0pt minus 2pt}
\titlespacing\subsection{0pt}{0pt plus 0pt minus 0pt}{0pt plus 0pt minus 2pt}
\titlespacing\subsubsection{0pt}{0pt plus 0pt minus 0pt}{0pt plus 0pt minus 2pt}

\newlength{\strutheight}
\newcommand{\head}[1]{\textnormal{\textbf{#1}}}

\usepackage{floatrow}
\usepackage{colortbl}

\newfloatcommand{capbtabbox}{table}[][\FBwidth]

\newcommand{\mygrayline}{\arrayrulecolor{gray}\hline}

\title{Feature Selection Based on Unique Relevant Information for Health Data \thanks{This work was supported in part by the National University of Singapore and the Singapore Ministry of Education under Grant R-263-000-D35-114.}}

\author{
  Shiyu Liu \normalfont{and} \textbf{Mehul Motani}\\
  Department of Electrical and Computer Engineering\\
  National University of Singapore\\
  \texttt{shiyu$\_$liu@u.nus.edu, motani@nus.edu.sg}  \\
}

\begin{document}
\maketitle
\vspace{-8mm}
\begin{abstract}
\vspace{-3mm}
Feature selection, which searches for the most representative features in observed data, is critical for health data analysis. Unlike feature extraction, such as PCA and autoencoder based methods, feature selection preserves interpretability, meaning that the selected features provide direct information about certain health conditions (i.e., the label). Thus, feature selection allows domain experts, such as clinicians, to understand the predictions made by machine learning based systems, as well as improve their own diagnostic skills. 
Mutual information is often used as a basis for feature selection since it measures dependencies between features and labels.
In this paper, we introduce a novel mutual information based feature selection (MIBFS) method called SURI, which boosts features with  high unique relevant information. 
We compare SURI to existing MIBFS methods using 3 different classifiers on 6 publicly available healthcare data sets.
The results indicate that, in addition to preserving interpretability, SURI selects more relevant feature subsets which lead to higher classification performance.  More importantly, we explore the dynamics of mutual information on a public low-dimensional health data set via exhaustive search. The results suggest the important role of unique relevant information in feature selection and verify the principles behind SURI.
\end{abstract}

\vspace{-1mm}
\section{Introduction}
Feature selection, which searches for the most representative features of the observed data, is critical for machine learning algorithms as different features can entangle and hide more or less the different explanatory factors of variation behind the data \cite{Bengio2013,Bermingham2015,Hoque2016,Zhou2017BCB,Jia2017BIBM}. More importantly, the feature selection process preserves the interpretability of raw data. Hence, the selected feature subset provides useful information about which features are indicative of certain health conditions \cite{Been2015}. Understanding these relationships is important for domain experts, such as clinicians, to understand machine learning based predictive diagnosis, as well as improve their own diagnostic skills. 

\begin{comment}
A natural metric to measure dependency between random variables is mutual information (MI) \cite{cover2006elements}. MI has a solid theoretical foundation in quantifying the dependency, in that MI can be used to write both an upper and lower bound on the Bayes error rate \cite{Fano1961, Brown}. Therefore, MI based feature selection (MIBFS) methods that attempted to maximize the MI between the selected features and the label are a type of popular classifier-independent feature selection methods \cite{Sun2014}.
A rich body of work in the past have been done to approximate the high-dimensional MI with low-dimensional MI terms \cite{Brown,Peng2005}. It is because computation of high-dimensional MI requires expensive integrations, which makes it computationally prohibitive when data dimension goes high. Although the idea of MIBFS seems trivial, existing MIBFS techniques proposed in the literature are still far from optimal especially in face of high-dimensional data.
\end{comment}

Mutual information (MI) measures the dependency between random variables \cite{cover2006elements} and is often used as a basis for feature selection.
In this paper, we propose a novel mutual information based feature selection (MIBFS) method called Selection via Unique Relevant Information (SURI). The key differences between SURI and existing MIBFS methods \cite{Meyer2008,Sun2014,Battiti,Lewis1992,Yang1999,Bennasar2015,Nguyen2014,Wu2016,Brown,Peng2005,Gao2016} are as follows: (i) SURI is the first to use the unique relevant information (URI) of each individual feature in feature selection. It searches for the optimal feature subset by boosting features with high URI; (ii) Different from existing MIBFS methods which approximate the high-dimensional joint MI with low-dimensional MI terms, SURI directly estimates the high-dimensional joint MI via nearest-neighbor based MI estimators \cite{Kraskov2003}.
%\begin{itemize}[label=$\diamond$,leftmargin=*] 
%	\item We present the first study to use the unique relevant information (URI) carried by each individual feature in feature selection. Our algorithm is to search for the optimal feature subset by boosting the features with high URI.
%	\item We directly estimate the high-dimensional joint MI via MI estimators instead of approximating the high-dimensional joint MI with low-dimensional MI terms.
%\end{itemize}

The main contributions of this paper are as follows:
\begin{itemize}[noitemsep,topsep=-4pt,label=$\bullet$,leftmargin=*]
\item We propose SURI to boost features with high Unique Relevant Information during the feature selection process in order to achieve better classification performance.
\item The proposed SURI algorithm outperforms other state-of-art MIBFS methods on 6 publicly available healthcare data sets for 3 classifiers. More importantly, the selected feature subset preserves the interpretability of raw data. Hence, it can provide useful information about which features are possibly indicative of certain health conditions.
\item Using a low-dimensional data set, we explore the dynamics of MI in feature selection via exhaustive search and establish the significance of URI. Based on experiments, we find that features with relatively large URI tend to be frequently present in the feature subsets with the highest classification accuracies (found via exhaustive search).
\end{itemize}

\section{Proposed Algorithm}
\subsection{Notation $\&$ Data Set Description}
Let \(\Omega\) denote the set of all features and let $n$ be the total number of features. Let $\mathcal{S}\subseteq \Omega$ be the set of selected features. Let $X\in\Omega$ be a feature and $Y$ be the classification target (label).  Let $H(X_i)$ denote the entropy of feature $X_i$ and $I(X_i;Y)$ denote the mutual information between feature $X_i$ and label $Y$ \cite{cover2006elements}. 

Six publicly available data sets, obtained from the UCI Machine Learning Repository, are used in our study.  The Z-Alizadeh Sani data set \cite{Ali}, Breast Cancer data set \cite{BC}, SPECTF data set \cite{SP}, Arrhythmia data set \cite{Arr} and EEG data set \cite{EEG} are used to examine the effectiveness of SURI and existing MIBFS methods. The Heart Disease data set \cite{HD} is used to study, via exhaustive search, the dynamics of MI and explore the role of URI. The EEG data set is preprocessed by calculating mean, maximum, minimum, standard deviation, maximum adjacent change and minimum adjacent change for each channel. The Heart Disease data set is aggregated into 2 different classes. Class 0 has 164 patients without heart disease, Class 1 combines 139 patients who have 4 different level of heart disease into one class. Information about each data set is presented in Table \ref{table_11}.

\subsection{Information Content: URI, ORI, II}

\begin{figure}
\begin{floatrow}
\resizebox{4.5in}{!}{
\ffigbox{%
  \includegraphics[height= 3.5cm,width= 6.3cm]{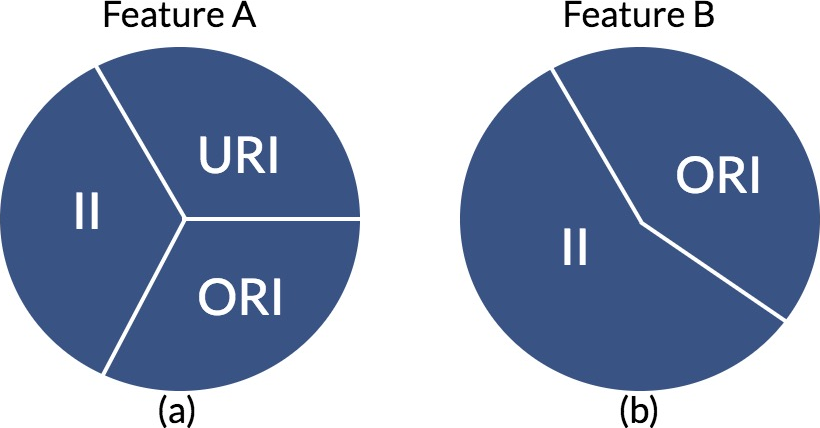}%
}{%

  \caption{Entropy of Feature A and Feature B from the perspective of URI,ORI,II.}%
  \label{chart}
}
\capbtabbox{%
  \begin{tabular}{ccccc} \hline
		\cmidrule{1-5}	
		& \head{Data Sets} & \head{Features}  & \head{Samples} & \head{\% of URI}\\ \cmidrule{1-5}
		\multicolumn{1}{c}{}    &   
		\multicolumn{1}{l}{Z-Alizadeh Sani}& 54 & 303 &11.1\%\\
		\multicolumn{1}{c}{}    &   
		\multicolumn{1}{l}{Breast Cancer} &30 & 569 &16.6\%\\
		\multicolumn{1}{c}{}    &
		\multicolumn{1}{l}{SPECTF Heart} &44 &267 &38.6\%\\
		\multicolumn{1}{c}{}    &
		\multicolumn{1}{l}{Arrhythmia} &279 & 452 &18.9\%\\
		\multicolumn{1}{c}{}    &
		\multicolumn{1}{l}{EEG}& 384 & 1200 &19.5\%\\
		\multicolumn{1}{c}{}    &
		\multicolumn{1}{l}{Heart Disease}& 13 & 303 &53.8\%\\
		\cmidrule{1-5}
  \end{tabular}
}{%
	\caption{Information of Health Data Sets Studied}%
	\label{table_11}
}
}
\end{floatrow}
\end{figure}

The information content of a feature (i.e., entropy of a feature) can be divided into two parts: relevant information (RI) and irrelevant information (II). Relevant information can be further divided into unique relevant information (URI) and overlapped relevant information (ORI).

Irrelevant information (II) can be understood as the noise in the signal which confuses the classifier and leads to lower accuracy \cite{John1994, Radha2016}. 

Relevant information can be understood as the non-zero MI between feature $X_i$ and the label $Y$ (i.e., $I(X_i;Y$)>0). URI is the unique relevant information content of a feature with the label which is not shared by any other features in $\Omega$. Mathematically, the URI of a feature is $I(\Omega;Y) - I(\Omega \backslash(X_{i});Y)$. This notion of URI is equivalent to strong relevance defined in \cite{KJ,Brown}. URI helps classifiers to differentiate labels and contributes to higher classification accuracy. In contrast to URI, ORI is the relevant information content of a feature with the label which is shared (or overlapped) with other features in $\Omega$ (i.e., ORI=RI-URI). This is equivalent to weak relevance defined in \cite{KJ,Brown}.

We illustrate the idea of URI and ORI in Figure \ref{chart}. Consider Feature A and Feature B shown in Figure \ref{chart}. Assume that Feature B does not contain significant URI ($I(\Omega;Y) - I(\Omega \backslash X_{B};Y) \to 0$). Thus, there exists a feature subset $\mathcal{P}\subset\Omega$ that overlaps with Feature B's total amount of relevant information. This means that the contribution of Feature B is a strict subset of $\mathcal{P}$'s contribution. 
Therefore, when $\mathcal{P}$ is selected, the ORI of Feature B will become redundant. Hence, selecting Feature B will not improve the classification performance (based on maximum relevance minimum redundancy criterion) \cite{Peng2005}, but include its undesired redundancy and irrelevance. On the other hand, Feature A contains non-negligible URI. So, even if Feature B (or any other feature) has been selected first, adding Feature A into the selected feature set can still improve the classification accuracy. 

\begin{table}[t]
	%\vspace{-20mm}	
	\captionsetup{font=footnotesize}
	\caption{Peak Accuracy (\# of features used), F1-Score, AUC-ROC of various algorithms on EEG data set via the RF classifier. Similar results can be observed for Alizadeh data set, Breast Cancer data set, SPECTF Heart data set and Arrhythmia data set via KNN, SVM, and RF. These results are shown in Appendix \ref{sec:app}.}
	\label{table_res}
	\resizebox{5in}{!}{
	\begin{tabular}{c c c c c c c c c}
		\cmidrule{1-8}
		&& \multicolumn{6}{c}{\head{Performance of Existing MIBFS Methods}}\\ \cmidrule{2-8}
		& GSA & MIM & JMI & (SPEC$_{CMI})$ & HOMIFS & JMIM &  \\ \cmidrule{2-8} 
		\multicolumn{1}{l}{\head{Peak Accuracy}}& \textbf{90.1\% (37)} & 87.1\% (47) & 87.4\% (43) & 87.5\% (42) & 88.1\% (51) & 87.6\% (49)  \\
		\multicolumn{1}{l}{\head{F1-Score}}& \textbf{0.902} & 0.873 & 0.871 & 0.874 &0.899 &0.891  \\
		\multicolumn{1}{l}{\head{AUC-ROC}}& \textbf{0.897} & 0.867 & 0.869 & 0.871 & 0.889  & 0.865 \\
		\cmidrule{1-8}
		&& \multicolumn{6}{c}{\head{Performance of GSA and SURI}}\\ \cmidrule{2-8}
		&& GSA & SURI($\beta$=0.1) & SURI($\beta$=0.2) & SURI($\beta$=0.3)& \\ \cmidrule{2-8} 
		\multicolumn{1}{l}{\head{Peak Accuracy}} && 90.1\% (37) & 90.7\% (27) & \textbf{91.2\% (33)} & 87.7\% (32) \\
		\multicolumn{1}{l}{\head{F1-Score}}&& 0.902 & 0.905 & \textbf{0.909} &0.893 \\
		\multicolumn{1}{l}{\head{AUC-ROC}}&& 0.897 &0.901 &\textbf{0.905} &0.887 \\
		\cmidrule{1-8}
	\end{tabular}
	}
\end{table}

For each health data set in Table \ref{table_11}, we computed the percentage of features that carry non-negligible ($>10^{-8}$) URI with respect to the label. It can be seen that, for many of the data sets, the percentage of features that carry non-negligible URI is quite small ($<20\%$). Therefore, it is likely that a feature containing no URI but high ORI will be selected first. To avoid this, features with high URI must be boosted during the feature selection process. This is the motivation behind SURI.

\subsection{Proposed Algorithm: SURI}
Our proposed feature selection algorithm, SURI, balances increasing relevancy and boosting URI. SURI assigns a score to each feature according to the scoring function given by
%To boost URI, maximizing relevance and minimizing redundancy in feature selection, we propose a novel MIBFS method, called Selection via Unique Relevant Information (SURI). Same as other Existing MIBFS algorithms, SURI assigns a score to each feature. Higher score means more important feature, then we include each feature into $\mathcal{S}$ according to its score. The scoring function of SURI is given by
\begin{equation} 
\label{eq-suri}
J_{SURI}(X) =  (1 - \beta)  I(\mathcal{S}, X;Y) +  \beta  (I(\Omega;Y) - I(\Omega \backslash X;Y)),
\end{equation}
where $\beta$ is a weighting parameter. The first term corresponds to increasing relevancy and the second term rewards features with URI. The URI term allows SURI to have an overview of all features and helps to calibrate feature selection. The URI of each feature needs to be calculated only once and can be used in subsequent computations. Hence, the time complexity of SURI is $O(n^2)$, which is comparable to many state-of-the art MIBFS methods (e.g., JMI \cite{Yang1999}, GSA \cite{Brown}).  Different from existing MIBFS methods, which approximate the high-dimensional joint MI via low-dimensional MI, SURI directly estimates the high-dimensional joint MI via nearest neighbor based approach \cite{Kraskov2003}. 

\section{Performance Comparison with Existing MIBFS Methods }
\label{sec3}
In this paper, we evaluate six representative MIBFS methods: Mutual Information Maximization (MIM) \cite{Lewis1992}, Joint Mutual Information (JMI) \cite{Yang1999}, Mutual Information Maximization via Spectral Relaxation (SPEC$_{CMI}$) \cite{Nguyen2014}, Joint Mutual Information Maximization (JMIM) \cite{Bennasar2015}, High Order Mutual Information Feature Selection (HOMIFS) \cite{Wu2016} and Greedy Search Algorithm (GSA) \cite{Brown}. We clarify that GSA selects the candidate feature with the largest joint MI with the label at each step. 
The results of different MIBFS methods on the EEG data set via the random forest (RF) classifier are shown in Table \ref{table_res}. We shortlist GSA as it has the highest accuracy among existing MIBFS methods and compare SURI to GSA in Table \ref{table_res}, for various values of $\beta$.
The results for other data sets in Table \ref{table_11} via RF, k-nearest neighbors (KNN) and support vector machine (SVM) are shown in Appendix \ref{sec:app}.
We observe that SURI has the highest performance over all existing methods studied.

\section{Exploring the Dynamics of MI and URI on Heart Disease Data Set}
\label{sec5}
In addition to the performance benefits of SURI, we also want to understand the dynamics of MI and the specific role of URI in performance. In this section, we explore a low-dimensional data set (Heart Disease data set \cite{HD}) via exhaustive search. We use the SVM classifier for this part of the study.

The different feature subsets obtained from exhaustive search are ranked based on their corresponding accuracies in descending order and the top 20 feature subsets are shown in Table \ref{table_1}. The optimal solution is  the set containing Features \{0, 1, 2, 7, 11, 12\} with accuracy of 0.8567.  The columns named HR(0) and HR(1) represent the hit rate of label 0 and label 1, respectively. 

Table \ref{table_2} shows some statistics derived from Table \ref{table_1} for each of the 13 features in the data set.  These include: (i) the URI of each feature measured by $I(\Omega;Y) - I(\Omega \backslash X ;Y)$, (ii) the frequency of each feature in the top 20 ranked feature sets, and (iii) the MI of the feature with the label, $I (X;Y)$. 

Using Table \ref{table_2}, we make the following observations which illustrate the dynamics of MI/URI and the working principles of SURI.
	\begin{itemize}[label=$\bullet$,leftmargin=*]
		 \begin{adjustbox}{valign=T,raise=\strutheight,minipage={1.0\linewidth}}
				\begin{wraptable}{c}{0.45\linewidth}
					\captionsetup{font=footnotesize}
					\caption{Top 20 feature subsets (found via exhaustive search) ranked by SVM accuracy}
					\label{table_1}
					\begin{small}
					\resizebox{2.4in}{!}{
					\begin{tabular}{llll}
					\hline
					Rank & Features (count) & Acc. &HR(0$\backslash$1) \\
					\hline
					1    & 0, 1, 2, 7, 11, 12 (6)                  & 0.8567 & 0.90$\backslash$0.80  \\
					2    & 2, 3, 6, 7, 8, 11, 12 (7)              & 0.8567&0.91$\backslash$0.79 \\
					3    & 2, 3, 4, 5, 6, 7, 8, 9, 11, 12 (10)   & 0.8560& 0.90$\backslash$0.79  \\
					4    & 2, 3, 4, 5, 7, 8, 9, 10, 11, 12 (10) & 0.8557& 0.89$\backslash$0.80        \\
					5    & 0, 1, 2, 4, 5, 7, 8, 9, 11, 12 (10)   & 0.8554&  0.88$\backslash$0.81 \\
					6    & 0, 2, 5, 6, 7, 8, 9, 10, 11, 12 (10) & 0.8550& 0.88$\backslash$0.81 \\
					7    & 2, 3, 5, 6, 7, 8, 11, 12 (8)& 0.8550&  0.89$\backslash$0.80 \\
					8    & 1, 2, 5, 7, 8, 9, 11, 12 (8)& 0.8547&  0.89$\backslash$0.81     \\
					9    & 0, 2, 3, 6, 8, 9, 11, 12 (8)& 0.8547&   0.88$\backslash$0.81      \\
					10  & 0, 2, 5, 6, 8, 9, 11, 12 (8)& 0.8540&  0.89$\backslash$0.80        \\
					11  & 0, 1, 2, 5, 6, 7, 8, 9, 11, 12 (10)& 0.8540& 0.91$\backslash$0.78     \\
					12  &1, 2, 3, 5, 6, 7, 8, 9, 11, 12 (10)& 0.8540& 0.89$\backslash$0.80       \\
					13  &0, 2, 3, 5, 6, 7, 8, 9, 11, 12 (10)& 0.8537& 0.90$\backslash$0.79        \\
					14  &2, 4, 6, 8, 9, 11, 12 (7)& 0.8537& 0.89$\backslash$0.80        \\
					15  & 0, 2, 6, 7, 8, 9, 10, 11, 12 (9)& 0.8534& 0.89$\backslash$0.80        \\
					16  & 0, 1, 2, 3, 4, 6, 7, 8, 10, 11, 12 (11)& 0.8531& 0.88$\backslash$0.80       \\
					17  &1, 2, 3, 4, 5, 6, 8, 9, 11, 12 (10)& 0.8525& 0.90$\backslash$0.79         \\
					18  &1, 2, 4, 6, 7, 8, 9, 11, 12 (9)& 0.8523& 0.88$\backslash$0.80        \\
					19  &0, 2, 3, 4, 5, 7, 8, 9, 10, 11, 12 (11)& 0.8523 &0.89$\backslash$0.79        \\
					20  &1, 2, 3, 6, 7, 8, 9, 10, 11, 12 (10)& 0.8523& 0.88$\backslash$0.80 \\
					\hline
				\end{tabular}
				}
			\end{small}	
		\end{wraptable}
\item Features with high MI tend to have non-zero URI as those features contain more relevant information. Those features (Features 2, 7, 8, 11, and 12) are more likely to have high frequencies in the top 20 ranking. This shows the correlation between MI and frequency and demonstrates that MI is a good criterion for feature selection. 

\item For features with similar MI, a feature with higher URI is more likely to have high frequency (Feature 7 \& Feature 10, Feature 8 \& Feature 9, Feature 1 \& Feature 6). The frequency of Feature 7 is much higher than that of Feature 10 even though they have almost the same MI with the label. We observe that Feature 7 contains URI while Feature 10 does not. This phenomenon also happens with the pair Feature 8 and  9 and the pair Feature 1 and 6. For Features 1, 4 and 10, even though Features 4 and 10 have a higher MI than Feature 1, the frequency of Feature 1 is the highest among those 3 features. We observe that Feature 1 contains URI while the URI of Features 4 and 10 is negligible ($<10^{-8}$). 

\end{adjustbox}

\begin{adjustbox}{valign=T,raise=\strutheight,minipage={1.0\linewidth}}
		\begin{wraptable}{c}{0.45\linewidth}
			\captionsetup{font=footnotesize}
			\caption{URI, Frequency, and MI of each feature in top 20 feature subsets}
			\label{table_2}
			\begin{small}
			\begin{sc}
			\resizebox{2.4in}{!}{
			\begin{tabular}{llll}
				\hline
				Feature & URI & Frequency & MI \\
				\hline
				0    & \textless \num{e-8}         & 10& 0.01303179 \\
				1    & 0.01864& 9  & 0.03936924 \\
				2    & 0.12847& 20& 0.14232146 \\
				3    & \textless \num{e-8}           & 11& 0.01091093        \\
				4    & \textless \num{e-8}          & 8  & 0.04722264\\
				5    & \textless \num{e-8}           &12 &  0.01303062\\
				6    & 0.03351& 15&  0.04370709       \\
				7    & 0.01432& 16&0.07347634        \\
				8    & 0.03884&19 & 0.09201738        \\
				9    & \textless \num{e-8}           &16 & 0.09474291        \\
				10  & \textless \num{e-8}           &6   &0.07714593        \\
				11  &0.11704  &20 & 0.12328436       \\
				12  & 0.07190&20&0.14631033 \\
				\hline
			\end{tabular}
			}
			\end{sc}
		\end{small}	
	\end{wraptable}	
\item Features with low MI and low URI tend to have a low frequency (e.g., Feature 	1). The frequency of Feature 1 is relatively low even though Feature 1 contains a small amount of URI. This is because low MI stands for high irrelevant information content, which is harmful to classification accuracy.

\item Features with negligible URI and high MI can also have high frequency (e.g., Feature 9). Since $X_{2},X_{8},X_{11},$ and $X_{12}$ appear in most subset of top 20 ranks, we calculate $I (X_{2},X_{8},X_{11},X_{12},X_{9} ; Y)$ and find that $I (X_{2},X_{8},X_{11},X_{12},X_{9} ; Y) > I(X_{2},X_{8},X_{11},X_{12}; Y)$, meaning that Feature 9 contributes to higher joint MI with respect to Features 2, 8, 11, \& 12, explaining the relatively high frequency of Feature 9.
 
\end{adjustbox}
\end{itemize}

\section{Reflections}

In the experimental results above, we see the working principle of SURI in action. The first term in \eqref{eq-suri}, $I(\mathcal{S}, X; Y)$, guarantees the relevance of the selected subset and the second term in \eqref{eq-suri}, $I(\Omega;Y) - I(\Omega \backslash X ;Y)$,  is to reward features with relatively higher URI. Therefore, features with high joint MI and URI will be selected earlier. For features with similar joint MI, the second term will prioritize features with higher URI. For features with high URI and low joint MI, the first term will penalize them due to their low joint MI.

%In addition, the results above also suggest that besides certain well-known features (Feature 2: Chest pain type; Feature 11: Number of major vessels by fluoroscopy; Feature 12: Thallium stress test result) \cite{Sab2014}, Feature 6 (Resting ECG result), Feature 7 (Maximum heart rate), Feature 8 (Exercise induced angina) and Feature 9 (ST depression induced by exercise relative to rest) also have a significant effect on heart disease diagnosis. Those features can be identified and selected by SURI to further study the diagnosis of heart disease.   
Table \ref{TDiscri} in Appendix \ref{sec:AppendixB} gives the physical description of each feature in the UCI Heart Disease data set. Referring to Tables \ref{table_2} and \ref{TDiscri}, our findings indicate that the following features have high frequency and should have an effect upon heart disease and its diagnosis: Feature 2 (chest pain type), Feature 6 (resting ECG result), Feature 7 (maximum heart rate), Feature 8 (exercise induced angina), Feature 9 (ST depression induced by exercise relative to rest), Feature 11 (number of major vessels colored by fluoroscopy) and Feature 12 (Thallium stress test result). The majority of these features have non-negligible URI and are more likely to be chosen by SURI. Additionally, the Thallium stress test is administered in patients with heart disease, which is consistent with the fact that Feature 12 has the highest MI and significant URI (see Table \ref{table_2}).

%Those features can be potentially identified and selected by SURI based on the working principle of SURI stated above. 

To see how far SURI is from optimal,  we compare it to the optimal solution and find that SURI is ranked 25th, which is highest amongst the MIBFS methods (see Figure \ref{fig:bar} in Appendix \ref{sec:AppendixB}). So while MI and URI lead to good suboptimal feature selection solutions, our results suggest that there are other hidden variables which could further improve the performance. We also have not specified how to select the $\beta$ parameter to optimally balance relevance and URI. We expect that these topics are worthy of investigation and will open up many interesting avenues for further exploration.

\newpage
\bibliographystyle{IEEEtran}
\bibliography{sample-bibliography}

\begin{thebibliography}{10}
\providecommand{\url}[1]{#1}
\csname url@samestyle\endcsname
\providecommand{\newblock}{\relax}
\providecommand{\bibinfo}[2]{#2}
\providecommand{\BIBentrySTDinterwordspacing}{\spaceskip=0pt\relax}
\providecommand{\BIBentryALTinterwordstretchfactor}{4}
\providecommand{\BIBentryALTinterwordspacing}{\spaceskip=\fontdimen2\font plus
\BIBentryALTinterwordstretchfactor\fontdimen3\font minus
  \fontdimen4\font\relax}
\providecommand{\BIBforeignlanguage}[2]{{%
\expandafter\ifx\csname l@#1\endcsname\relax
\typeout{** WARNING: IEEEtran.bst: No hyphenation pattern has been}%
\typeout{** loaded for the language `#1'. Using the pattern for}%
\typeout{** the default language instead.}%
\else
\language=\csname l@#1\endcsname
\fi
#2}}
\providecommand{\BIBdecl}{\relax}
\BIBdecl

\bibitem{Bengio2013}
Y.~Bengio, A.~Courville, and P.~Vincent, ``Representation learning: A review
  and new perspectives,'' \emph{IEEE Trans. Pattern Anal. Mach. Intell.},
  vol.~35, no.~8, pp. 1798--1828, Aug 2013.

\bibitem{Bermingham2015}
M.~Bermingham \emph{et~al.}, ``Application of high-dimensional feature
  selection: evaluation for genomic prediction in man,'' \emph{Scientific
  Reports}, vol.~5, no. 10312, 2015.

\bibitem{Hoque2016}
H.~Hoque, N.and~Ahmed, D.~Bhattacharyya, and J.~Kalita, ``A fuzzy mutual
  information-based feature selection method for classification,'' \emph{Fuzzy
  Information and Engineering}, vol.~8, no.~3, pp. 355--384, 2016.

\bibitem{Zhou2017BCB}
C.~Zhou, J.~Yao, M.~Motani, and J.~Chew, ``Learning deep representations from
  heterogeneous patient data for predictive diagnosis,'' in \emph{ACM
  International Conference on Bioinformatics, Computational Biology,and Health
  Informatics (ACM-BCB)}, Aug. 2017, pp. 115--123.

\bibitem{Jia2017BIBM}
J.~Yao, C.~Zhou, and M.~Motani, ``Spatio-temporal autoencoder for feature
  learning in patient data with missing observations,'' in \emph{IEEE
  International Conference on Bioinformatics and Biomedicine (BIBM)}, Nov.
  2017, pp. 886--890.

\bibitem{Been2015}
B.~Kim, J.~Shah, and F.~Doshi-Velez, ``Mind the gap: A generative approach to
  interpretable feature selection and extraction,'' \emph{Conference on Neural
  Information Processing Systems (NIPS)}, vol. 280, no.~19, pp. 1690--1691,
  2015.

\bibitem{cover2006elements}
T.~M. Cover and J.~A. Thomas, \emph{Elements of Information Theory, 2nd
  edition}.\hskip 1em plus 0.5em minus 0.4em\relax John Wiley \& Sons, 2006.

\bibitem{Meyer2008}
P.~Meyer, C.~Schretter, and G.~Bontempi, ``Information-theoretic feature
  selection in microarray data using variable complementarity,'' \emph{IEEE J.
  Sel. Topics Signal Process.}, vol.~2, pp. 261 -- 274, 2008.

\bibitem{Sun2014}
L.~Sun and J.~Xu, ``Feature selection using mutual information based
  uncertainty measures for tumor classification,'' \emph{Bio-Medical Materials
  and Engineering}, vol.~24, no.~1, pp. 763--770, 2014.

\bibitem{Battiti}
R.~Battiti, ``Using mutual information for selecting features in supervised
  neural net learning,'' \emph{IEEE Trans. Neural Netw.}, vol.~5, no.~4, 1994.

\bibitem{Lewis1992}
D.~Lewis, ``Feature selection and feature extraction for text categorization,''
  \emph{Proceedings of the workshop on Speech and Natural Language}, pp. 212 --
  217, 1992.

\bibitem{Yang1999}
H.~Yang and J.~Moody, ``Data visualisation and feature selection: New algorithm
  for nongaussian data,'' \emph{Conference on Neural Information Processing
  Systems (NIPS)}, vol.~99, pp. 687--693, 1999.

\bibitem{Bennasar2015}
M.~Bennasar, Y.~Hicks, and R.~Setchi, ``Feature selection using joint mutual
  information maximisation,'' \emph{Expert System with Applications}, vol.~42,
  no.~22, pp. 8520 -- 8532, 2015.

\bibitem{Nguyen2014}
X.~Nguyen, J.~Chan, S.~Romano, and J.~Bailey, ``Effective global approaches for
  mutual information based feature selection,'' \emph{ACM SIGKDD international
  conference on Knowledge discovery and data mining}, pp. 512 -- 521, Aug.
  2014.

\bibitem{Wu2016}
J.~Wu, S.~Gupta, and C.~Bajaj, ``High order mutual information approximation
  for feature selection,'' \emph{arXiv preprint arXiv:1612.00554}, 2016.

\bibitem{Brown}
G.~Brown, A.~Pancock, M.~Zhao, and M.~Lujan, ``Conditional likelihood
  maximisation: A unifying framework for information theoretic feature
  selection,'' \emph{J. Mach. Learn. Res.}, no.~13, pp. 27--66, 2012.

\bibitem{Peng2005}
H.~Peng, F.~Long, and C.~Ding, ``Feature selection based on mutual information:
  criteria of max-dependency, max-relevance, and min-redundancy,'' \emph{IEEE
  Trans. Pattern Anal. Mach. Intell.}, vol.~27, pp. 1226--1238, 2005.

\bibitem{Gao2016}
S.~Gao, G.~Steeg, and A.~Galstyan, ``Variational information maximization for
  feature selection,'' \emph{arXiv preprint arXiv:1606.02827}, 2016.

\bibitem{Kraskov2003}
A.~Kraskov, H.~St{\"o}gbauer, and P.~Grassberger, ``Estimating mutual
  information,'' \emph{Physical Review E69, 066138}, 2004.

\bibitem{Ali}
{UCI Alizadeh Data}, 2018,
  \url{https://archive.ics.uci.edu/ml/datasets/Z-Alizadeh+Sani}.

\bibitem{BC}
{UCI Breast Cancer Data}, 2018,
  \url{https://archive.ics.uci.edu/ml/datasets/Breast+Cancer+Wisconsin+(Diagnostic)}.

\bibitem{SP}
{UCI SPECTF Heart Data}, 2018,
  \url{http://archive.ics.uci.edu/ml/datasets/SPECTF+Heart}.

\bibitem{Arr}
{UCI Arrhythmia Data}, 2018,
  \url{https://archive.ics.uci.edu/ml/datasets/arrhythmia}.

\bibitem{EEG}
{UCI EEG Data}, 2018,
  \url{https://archive.ics.uci.edu/ml/datasets/eeg+database}.

\bibitem{HD}
{UCI Heart Disease Data}, 2018,
  \url{http://archive.ics.uci.edu/ml/datasets/heart+Disease}.

\bibitem{John1994}
G.~John, R.~Kohavi, and K.~Pfleger, ``Irrelevant features and the subset
  selection problem,'' \emph{International Conference on Machine Learning}, pp.
  121--129, Jul. 1994.

\bibitem{Radha2016}
Radha{\ }R.{\ }and{\ }Muralidhara{\ }S., ``Removal of redundant and irrelevant
  data from training datasets using speedy feature selection method,''
  \emph{Int'l J Comp. Sci. and Mob. Comput.}, vol.~5, no.~7, pp. 359--364,
  2016.

\bibitem{KJ}
R.~Kohavi and G.~John, ``Wrappers for feature subset selection,''
  \emph{Artificial Intelligence 97}, pp. 273--324, 1997.

\end{thebibliography}

\appendix
\section{Supplementary Results for Section \ref{sec3}}
\label{sec:app}

Tables \ref{T5} - \ref{T8} show the performance metrics (peak accuracy, AUC-ROC and F1-score) for existing MIBFS algorithms and SURI (for different $\beta$ values) on five data sets using the KNN classifier.

Tables \ref{T9} - \ref{T12} show the performance metrics (peak accuracy, AUC-ROC and F1-score) for existing MIBFS algorithms and SURI (for different $\beta$ values) on five data sets using the SVM classifier.

Tables \ref{T13} - \ref{T16} show the performance metrics (peak accuracy, AUC-ROC and F1-score) for existing MIBFS algorithms and SURI (for different $\beta$ values) on five data sets using the Random Forest classifier.

\begin{table}[!htb] 
	\caption{Peak Accuracy of Existing MIBFS Algorithms using KNN.}
	\label{T5}
	%\begin{minipage}[b]{0.45\linewidth}\centering
	\resizebox{5in}{!}{
	\begin{tabularx}{\linewidth}{c c c c c c c c}
		\cmidrule{1-8}
		&& \multicolumn{6}{c}{\head{Peak Accuracy (No. of Features Used)}}\\ \cmidrule{3-8}
		%                & & \multicolumn{5}{c}{\head{Data Set Information$\&$Peak Accuracy (Number of Features Used)}}\\ \cmidrule{3-11}
		&& GSA & MIM & JMI & SPEC$_{CMI}$ & HOMIFS & JMIM  \\ \cmidrule{3-8} 
		\multicolumn{1}{c}{\multirow{5}{*}{\begin{sideways}\parbox{2cm}{\centering Data Set \\ Name}\end{sideways}}}   &
		\multicolumn{1}{l}{Alizadeh}& \textbf{81.6\% (25)} & 77.7\% (30) & 80.3\% (28) & 80.6\% (29) & 79.0\% (31) & 80.3\% (27)  \\
		\multicolumn{1}{c}{}    &   
		\multicolumn{1}{l}{Breast Cancer} & \textbf{92.7\% (6)} & 91.9\% (8) & 90.9\% (9) & 92.3\% (7) & 91.0\% (10) & 90.7\% (7)  \\
		\multicolumn{1}{c}{}    &
		\multicolumn{1}{l}{SPECTF Heart} & \textbf{60.8\% (14)} & 57.9\% (16) & 59.1\% (13) & 60.5\% (15) & 58.8\% (14) & 59.3\% (15) \\
		\multicolumn{1}{c}{}    &   
		\multicolumn{1}{l}{Arrhythmia} & \textbf{72.5\% (33)} & 69.4\% (37) & 72.3\% (34) & 68.1\% (35) & 68.7\% (38) & 67.4\% (31)\\
		\multicolumn{1}{c}{}    &   
		\multicolumn{1}{l}{EEG}& \textbf{90.3\% (27)} & 86.1\% (32) & 87.3\% (27) & 88.1\% (31) & 89.1\% (27) & 88.3\% (51)  \\
		\cmidrule{1-8}
	\end{tabularx}
	}
	\hspace{11mm}
\end{table}

\begin{table}[h!]	
	\centering
	\setlength\tabcolsep{1.5pt} % default value: 6pt
	\caption[]{AUC-ROC $\&$ F1-Score Corresponding to the Peak Accuracy of Existing MIBFS Algorithms using KNN.}
		\label{T6}
	\resizebox{5in}{!}{
	\begin{tabularx}{\linewidth}{c c c c c c c c c c c c c c c}
		\cmidrule{1-15}
		&& \multicolumn{5}{c}{\head{AUC-ROC}} & \multicolumn{8}{c}{\head{F1-Score}}\\ \cmidrule{3-8} \cmidrule{10-15} 
		%                & & \multicolumn{5}{c}{AUC-ROC} {F1-Score} \\ \cmidrule{3-8} \cmidrule{10-15} 
		&& GSA  & MIM & JMI & SPEC$_{CMI}$ & HOMIFS & JMIM & & GSA & MIM & JMI & SPEC$_{CMI}$ &HOMIFS & JMIM \\ \cmidrule{3-8} \cmidrule{10-15}
		\multicolumn{1}{c}{\multirow{5}{*}{\begin{sideways}\parbox{2cm}{\centering Data Set \\ Name}\end{sideways}}}   &
		\multicolumn{1}{l}{Alizadeh}& \textbf{0.821} & 0.799 & 0.805 & 0.813 & 0.802  & 0.809 & & \textbf{0.814} & 0.781 & 0.801 & 0.811 &0.792 &0.799  \\
		\multicolumn{1}{c}{}    &
		\multicolumn{1}{l}{Breast Cancer} &\textbf{0.931} &0.922 & 0.907 & 0.926 & 0.915 & 0.911 & &\textbf{0.923} & 0.917 & 0.907 &0.913 &0.913 &0.909   \\
		\multicolumn{1}{c}{}    &
		\multicolumn{1}{l}{SPECTF Heart} &\textbf{0.613} &0.581 &0.588&0.603 & 0.591 & 0.595 & &\textbf{0.609} & 0.588 & 0.583 &0.599 &0.593&0.591 \\
		\multicolumn{1}{c}{}    &
		\multicolumn{1}{l}{Arrhythmia} & 0.727 & 0.707 & \textbf{0.729} & 0.677 & 0.681 & 0.679 & &0.721 & 0.701 & \textbf{0.723} &0.681&0.688&0.675\\
		\multicolumn{1}{c}{}    &   
		\multicolumn{1}{l}{EEG}& \textbf{0.899} & 0.857 & 0.861 & 0.878 & 0.893  & 0.871 & & \textbf{0.901} & 0.865 & 0.871 & 0.875 &0.883 &0.875  \\
		\cmidrule{1-15}
	\end{tabularx}
	}
	\hspace{11mm}
\end{table}

\begin{table}[!htb]
	\centering
	\setlength\tabcolsep{9pt} % default value: 6pt
	\caption{Peak Accuracy of SURI and GSA using KNN.}
		\label{T7}
	\resizebox{5in}{!}{
	\begin{tabular*}{\linewidth}{c c c c c c c}
		\cmidrule{1-6}
		&& \multicolumn{4}{c}{\head{Peak Accuracy (No. of Features Used)}}\\ \cmidrule{3-6} 
		&& GSA & SURI($\beta$=0.1) & SURI($\beta$=0.2) & SURI($\beta$=0.3) \\ \cmidrule{3-6} 
		\multicolumn{1}{c}{\multirow{5}{*}{\begin{sideways}\parbox{2cm}{\centering Data Set \\ Name}\end{sideways}}}   &
		\multicolumn{1}{l}{Alizadeh}& 81.6\% (25) & 82.7\% (26) & \textbf{83.1\% (26)} & 78.6\% (27)   \\
		\multicolumn{1}{c}{}    &
		\multicolumn{1}{l}{Breast Cancer} & 92.7\% (6) & 93.1\% (7) & \textbf{93.8\% (7)} & 90.7\% (8)  \\
		\multicolumn{1}{c}{}    &
		\multicolumn{1}{l}{SPECTF Heart} & 60.8\% (14) & 61.5\% (15) & \textbf{61.8\% (15)} & 60.9\% (15) \\
		\multicolumn{1}{c}{}    &              
		\multicolumn{1}{l}{Arrhythmia} & 72.5\% (33) & \textbf{74.7\% (34)} & 73.9\% (33) & 71.1\% (35) \\
		\multicolumn{1}{c}{}    &
		\multicolumn{1}{l}{EEG} & 90.3\% (27) & \textbf{90.5\% (37)} & 89.7\% (33) & 88.7\% (19) \\
		\cmidrule{1-6}
	\end{tabular*}
	}
	\hspace{11mm}
\end{table}

\begin{table}[!htb]
	\centering
	\setlength\tabcolsep{1.5pt} % default value: 6pt
	\caption{AUC-ROC $\&$ F1-Score Corresponding to the Peak Accuracy of SURI and GSA using KNN.}
		\label{T8}
	\resizebox{5in}{!}{
	\begin{tabular*}{\linewidth}{c c c c c c c c c c c}
		\cmidrule{1-11}
		&& \multicolumn{3}{c}{\head{AUC-ROC}} & \multicolumn{6}{c}{\head{F1-Score}}\\ \cmidrule{3-6}  \cmidrule{8-11} 
		%                & & \multicolumn{5}{c}{AUC-ROC} {F1-Score} \\ \cmidrule{3-8} \cmidrule{10-15} 
		&& GSA & SURI($\beta$=0.1) & SURI($\beta$=0.2) & SURI($\beta$=0.3)& &GSA & SURI($\beta$=0.1) & SURI($\beta$=0.2) & SURI($\beta$=0.3) \\ \cmidrule{3-6} \cmidrule{8-11}
		\multicolumn{1}{c}{\multirow{5}{*}{\begin{sideways}\parbox{2cm}{\centering Data Set \\ Name}\end{sideways}}}   &
		\multicolumn{1}{l}{Alizadeh}& 0.821 & 0.827 & \textbf{0.829} & 0.803 & & 0.814 & 0.817 & \textbf{0.818} & 0.791  \\
		\multicolumn{1}{c}{}    &
		\multicolumn{1}{l}{Breast Cancer} &0.931 &0.933 & \textbf{0.934} & 0.922 & &0.923 & 0.922 & \textbf{0.934} &0.919  \\
		\multicolumn{1}{c}{}    &
		\multicolumn{1}{l}{SPECTF Heart} &0.613 &0.617 &\textbf{0.619} &0.606 & &0.609 & 0.614 & \textbf{0.615} &0.608 \\
		\multicolumn{1}{c}{}    & 
		\multicolumn{1}{l}{Arrhythmia} & 0.727 & \textbf{0.743} & 0.741 & 0.709& &0.721 & \textbf{0.741} & 0.733 &0.711\\
		\multicolumn{1}{c}{}    &
		\multicolumn{1}{l}{EEG} &\textbf{0.899} &0.898 &0.887 &0.881 & &\textbf{0.901} & 0.898 & 0.895 &0.893 \\
		\cmidrule{1-11}
	\end{tabular*}
	}
	\hspace{11mm}
\end{table}

%\subsection{performance and Results with SVM}

\begin{table}[!htb]
	\centering
	\setlength\tabcolsep{9pt} % default value: 6pt
	\caption{Peak Accuracy of Existing MIBFS Algorithms using SVM.}
		\label{T9}
	\resizebox{5in}{!}{
	\begin{tabular*}{\linewidth}{c c c c c c c c c}
		\cmidrule{1-8}
		&& \multicolumn{6}{c}{\head{Peak Accuracy (No. of Features Used)}}\\ \cmidrule{3-8}
		%                & & \multicolumn{5}{c}{\head{Data Set Information$\&$Peak Accuracy (Number of Features Used)}}\\ \cmidrule{3-11}
		&& GSA & MIM & JMI & SPEC$_{CMI}$ & HOMIFS & JMIM  \\ \cmidrule{3-8} 
		\multicolumn{1}{c}{\multirow{5}{*}{\begin{sideways}\parbox{2cm}{\centering Data Set \\ Name}\end{sideways}}}   &
		\multicolumn{1}{l}{Alizadeh}& \textbf{83.2\% (6)} & 81.3\% (7) & 81.9\% (9) & 82.8\% (10) & 82.9\% (8) & 80.5\% (9)  \\
		\multicolumn{1}{c}{}    &   
		\multicolumn{1}{l}{Breast Cancer} & \textbf{85.8\% (4)} & 83.1\% (2) & 82.4\% (3) & 85.5\% (5) & 83.4\% (2) & 85.5\% (3)  \\
		\multicolumn{1}{c}{}    &
		\multicolumn{1}{l}{SPECTF Heart} & \textbf{61.9\% (7)} & 60.3\% (9) & 58.8\% (9) & 60.2\% (8) & 60.3\% (10) & 59.3\% (11) \\
		\multicolumn{1}{c}{}    &   
		\multicolumn{1}{l}{Arrhythmia} & \textbf{62.0\% (25)} & 58.2\% (27) & 60.0\% (29) & 58.6\% (28) & 60.6\% (26) & 58.9\% (27)\\
		\multicolumn{1}{c}{}    &   
		\multicolumn{1}{l}{EEG}& \textbf{83.3\% (49)} & 78.5\% (22) & 81.7\% (37) & 80.3\% (41) & 81.9\% (46) & 82.6\% (52)  \\
		\cmidrule{1-8}
	\end{tabular*}	
	}
	\hspace{11mm}
\end{table}

\begin{table}[!htb]
	\centering
	\setlength\tabcolsep{1.5pt} % default value: 6pt
	\caption{AUC-ROC $\&$ F1-Score Corresponding to the Peak Accuracy of Existing MIBFS Algorithms using SVM.}
		\label{T10}
	\resizebox{5in}{!}{
	\begin{tabular*}{\linewidth}{c c c c c c c c c c c c c c c}
		\cmidrule{1-15}
		&& \multicolumn{5}{c}{\head{AUC-ROC}} & \multicolumn{8}{c}{\head{F1-Score}}\\ \cmidrule{3-8} \cmidrule{10-15} 
		%                & & \multicolumn{5}{c}{AUC-ROC} {F1-Score} \\ \cmidrule{3-8} \cmidrule{10-15} 
		&& GSA  & MIM & JMI & SPEC$_{CMI}$ & HOMIFS & JMIM & & GSA & MIM & JMI & SPEC$_{CMI}$ &HOMIFS & JMIM \\ \cmidrule{3-8} \cmidrule{10-15}
		\multicolumn{1}{c}{\multirow{5}{*}{\begin{sideways}\parbox{2cm}{\centering Data Set \\ Name}\end{sideways}}}   &
		\multicolumn{1}{l}{Alizadeh}& \textbf{0.836} & 0.817 & 0.821 & 0.833 & 0.829  & 0.808 & & \textbf{0.833} & 0.823 & 0.816 & 0.823 &0.827 &0.801  \\
		\multicolumn{1}{c}{}    &
		\multicolumn{1}{l}{Breast Cancer} &\textbf{0.861} &0.843 & 0.832 & 0.857 & 0.836 & 0.856 & &\textbf{0.857} & 0.833 & 0.824 &0.843 &0.839 &0.847   \\
		\multicolumn{1}{c}{}    &
		\multicolumn{1}{l}{SPECTF Heart} &\textbf{0.621} &0.607 &0.591&0.607 & 0.606 & 0.595 & &\textbf{0.615} & 0.609 & 0.588 &0.605 &0.607&0.593 \\
		\multicolumn{1}{c}{}    &
		\multicolumn{1}{l}{Arrhythmia} & \textbf{0.625} & 0.591 & 0.611 & 0.607 & 0.619 & 0.587 & &\textbf{0.622} & 0.595 & 0.607 &0.609&0.611&0.598\\
		\multicolumn{1}{c}{}    &   
		\multicolumn{1}{l}{EEG}& \textbf{0.835} & 0.791 & 0.821 & 0.809 & 0.825  & 0.831 & & \textbf{0.831} & 0.784 & 0.813 & 0.801 &0.817 &0.827  \\
		\cmidrule{1-15}
	\end{tabular*}
	}
	\hspace{11mm}
\end{table}

\begin{table}[!htb]
	\centering
	\setlength\tabcolsep{9pt} % default value: 6pt
	\caption{Peak Accuracy of SURI and GSA using SVM.}
		\label{T11}
	\resizebox{5in}{!}{
	\begin{tabular*}{\linewidth}{c c c c c c c}
		\cmidrule{1-6}
		&& \multicolumn{4}{c}{\head{Peak Accuracy (No. of Features Used)}}\\ \cmidrule{3-6} 
		&& GSA & SURI($\beta$=0.1) & SURI($\beta$=0.2) & SURI($\beta$=0.3) \\ \cmidrule{3-6} 
		\multicolumn{1}{c}{\multirow{5}{*}{\begin{sideways}\parbox{2cm}{\centering Data Set \\ Name}\end{sideways}}}   &
		\multicolumn{1}{l}{Alizadeh}& 83.2\% (6) & 82.7\% (7) & \textbf{83.9\% (7)} & 81.9\% (8)   \\
		\multicolumn{1}{c}{}    &
		\multicolumn{1}{l}{Breast Cancer} & 85.8\% (4) & 84.4\% (4) & \textbf{86.2\% (4)} & 84.9\% (5)  \\
		\multicolumn{1}{c}{}    &
		\multicolumn{1}{l}{SPECTF Heart} & 61.9\% (7) & 60.1\% (8) & 62.3\% (10) & \textbf{63.3\% (9)} \\
		\multicolumn{1}{c}{}    &              
		\multicolumn{1}{l}{Arrhythmia} & 62.0\% (25) & 62.7\% (27) & 63.1\% (28) & \textbf{63.3\% (28)} \\
		\multicolumn{1}{c}{}    &
		\multicolumn{1}{l}{EEG} & 83.3\% (49) & 84.1\% (33) & 83.1\% (29) & \textbf{84.9\% (42)} \\
		\cmidrule{1-6}
	\end{tabular*}
	}
	\hspace{11mm}
\end{table}

\begin{table}[!htb]
	\centering
	\setlength\tabcolsep{1.5pt} % default value: 6pt
	\caption{AUC-ROC $\&$ F1-Score Corresponding to the Peak Accuracy of SURI and GSA using SVM.}
			\label{T12}
	\resizebox{5in}{!}{
	\begin{tabular*}{\linewidth}{c c c c c c c c c c c}
		\cmidrule{1-11}
		&& \multicolumn{3}{c}{\head{AUC-ROC}} & \multicolumn{6}{c}{\head{F1-Score}}\\ \cmidrule{3-6}  \cmidrule{8-11} 
		%                & & \multicolumn{5}{c}{AUC-ROC} {F1-Score} \\ \cmidrule{3-8} \cmidrule{10-15} 
		&& GSA & SURI($\beta$=0.1) & SURI($\beta$=0.2) & SURI($\beta$=0.3)& &GSA & SURI($\beta$=0.1) & SURI($\beta$=0.2) & SURI($\beta$=0.3) \\ \cmidrule{3-6} \cmidrule{8-11}
		\multicolumn{1}{c}{\multirow{5}{*}{\begin{sideways}\parbox{2cm}{\centering Data Set \\ Name}\end{sideways}}}   &
		\multicolumn{1}{l}{Alizadeh}& 0.836 & 0.833 & \textbf{0.845} & 0.827 & & 0.833 & 0.825 & \textbf{0.836} & 0.807  \\
		\multicolumn{1}{c}{}    &
		\multicolumn{1}{l}{Breast Cancer} &0.861 &0.843 & \textbf{0.863} & 0.851 & &0.857 & 0.843 & \textbf{0.859} &0.853  \\
		\multicolumn{1}{c}{}    &
		\multicolumn{1}{l}{SPECTF Heart} &0.621 &0.603 & 0.622 & \textbf{0.631} & &0.615 & 0.605 & 0.619 &\textbf{0.621} \\
		\multicolumn{1}{c}{}    & 
		\multicolumn{1}{l}{Arrhythmia} & 0.625 & 0.617 & 0.629 & \textbf{0.630}& &0.622 & 0.619 & 0.623 &\textbf{0.624}\\
		\multicolumn{1}{c}{}    &
		\multicolumn{1}{l}{EEG} &0.835 &0.847 &0.838 &\textbf{0.851} & &0.831 & 0.835 & 0.827 &\textbf{0.841} \\
		\cmidrule{1-11}
	\end{tabular*}
	}
	\hspace{11mm}
\end{table}

\begin{table*}[!htb]
	\centering
	\caption{Peak Accuracy of Existing MIBFS Algorithms using Random Forest.}
			\label{T13}
	\label{table_9}
	\resizebox{5in}{!}{
	\begin{tabular*}{\linewidth}{c c c c c c c c c}
		\cmidrule{1-8}
		&& \multicolumn{6}{c}{\head{Peak Accuracy (No. of Features Used)}}\\ \cmidrule{3-8}
		%                & & \multicolumn{5}{c}{\head{Data Set Information$\&$Peak Accuracy (Number of Features Used)}}\\ \cmidrule{3-11}
		&& GSA & MIM & JMI & SPEC$_{CMI}$ & HOMIFS & JMIM  \\ \cmidrule{3-8} 
		\multicolumn{1}{c}{\multirow{5}{*}{\begin{sideways}\parbox{2cm}{\centering Data Set \\ Name}\end{sideways}}}   &
		\multicolumn{1}{l}{Alizadeh}& \textbf{85.4\% (28)} & 84.9\% (31) & 85.1\% (29) & 84.4\% (31) & 85.1\% (33) & 84.8\% (30)  \\
		\multicolumn{1}{c}{}    &   
		\multicolumn{1}{l}{Breast Cancer} & \textbf{95.8\% (6)} & 94.1\% (9) & 95.3\% (8) & 94.4\% (14) & 95.1\% (13) & 94.8\% (12)  \\
		\multicolumn{1}{c}{}    &
		\multicolumn{1}{l}{SPECTF Heart} & \textbf{64.3\% (14)} & 61.7\% (17) & 63.3\% (15) & 62.7\% (16) & 64.0\% (18) & 63.3\% (16) \\
		\multicolumn{1}{c}{}    &   
		\multicolumn{1}{l}{Arrhythmia} & \textbf{81.3\% (39)} & 79.2\% (51) & 80.8\% (46) & 80.4\% (45) & 80.2\% (55) & 78.6\% (43)\\
		\multicolumn{1}{c}{}    &   
		\multicolumn{1}{l}{EEG}& \textbf{90.1\% (37)} & 87.1\% (47) & 87.4\% (43) & 87.5\% (42) & 88.1\% (51) & 87.6\% (49)  \\
		\cmidrule{1-8}
	\end{tabular*}
	}
	\hspace{11mm}
\end{table*}

\begin{table*}[!htb]	
	\centering
	\setlength\tabcolsep{1.5pt} % default value: 6pt
	\caption{AUC-ROC $\&$ F1-Score Corresponding to the Peak Accuracy of Existing MIBFS Algorithms using Random Forest}
			\label{T14}
	\label{table_4}
	\resizebox{5in}{!}{
	\begin{tabular*}{\linewidth}{c c c c c c c c c c c c c c c}
		\cmidrule{1-15}
		&& \multicolumn{5}{c}{\head{AUC-ROC}} & \multicolumn{8}{c}{\head{F1-Score}}\\ \cmidrule{3-8} \cmidrule{10-15} 
		%                & & \multicolumn{5}{c}{AUC-ROC} {F1-Score} \\ \cmidrule{3-8} \cmidrule{10-15} 
		&& GSA  & MIM & JMI & SPEC$_{CMI}$ & HOMIFS & JMIM & & GSA & MIM & JMI & SPEC$_{CMI}$ &HOMIFS & JMIM \\ \cmidrule{3-8} \cmidrule{10-15}
		\multicolumn{1}{c}{\multirow{5}{*}{\begin{sideways}\parbox{2cm}{\centering Data Set \\ Name}\end{sideways}}}   &
		\multicolumn{1}{l}{Alizadeh}& \textbf{0.806} & 0.801 & 0.802 & 0.800 & 0.802  & 0.797 & & \textbf{0.901} & 0.893 & 0.890 & 0.894 &0.898 &0.886  \\
		\multicolumn{1}{c}{}    &
		\multicolumn{1}{l}{Breast Cancer} &\textbf{0.945} &0.934 & 0.940 & 0.943 & 0.927 & 0.938 & &\textbf{0.938} & 0.922 & 0.923 &0.925 &0.920 &0.913   \\
		\multicolumn{1}{c}{}    &
		\multicolumn{1}{l}{SPECTF Heart} &0.612 &0.607 &0.604&0.595 & \textbf{0.625} & 0.602 & &0.723 & 0.710 & 0.727 &0.714 &0.707&\textbf{0.729} \\
		\multicolumn{1}{c}{}    &
		\multicolumn{1}{l}{Arrhythmia} & \textbf{0.765} & 0.703 & 0.733 & 0.752 & 0.715 & 0.744 & &\textbf{0.742} & 0.711 & 0.763 &0.740&0.732&0.715\\
		\multicolumn{1}{c}{}    &   
		\multicolumn{1}{l}{EEG}& \textbf{0.897} & 0.867 & 0.869 & 0.871 & 0.889  & 0.865 & & \textbf{0.902} & 0.873 & 0.871 & 0.874 &0.899 &0.891  \\
		\cmidrule{1-15}
	\end{tabular*}
	}
	\hspace{11mm}
\end{table*}

\begin{table*}[!htb]
	\centering
	\setlength\tabcolsep{9pt} % default value: 6pt
	\caption{Peak Accuracy of SURI and GSA using Random Forest. }
			\label{T15}
	\label{table_suri}
	\resizebox{5in}{!}{
	\begin{tabular*}{\linewidth}{c c c c c c c}
		\cmidrule{1-6}
		&& \multicolumn{4}{c}{\head{Peak Accuracy (No. of Features Used)}}\\ \cmidrule{3-6} 
		&& GSA & SURI($\beta$=0.1) & SURI($\beta$=0.2) & SURI($\beta$=0.3) \\ \cmidrule{3-6} 
		\multicolumn{1}{c}{\multirow{5}{*}{\begin{sideways}\parbox{2cm}{\centering Data Set \\ Name}\end{sideways}}}   &
		\multicolumn{1}{l}{Alizadeh}& 85.4\% (28) & 85.9\% (32) & \textbf{86.3\% (33)} & 85.3\% (33)   \\
		\multicolumn{1}{c}{}    &
		\multicolumn{1}{l}{Breast Cancer} & 95.8\% (6) & 96.1\% (8) & \textbf{96.8\% (9)} & 95.0\% (7)  \\
		\multicolumn{1}{c}{}    &
		\multicolumn{1}{l}{SPECTF Heart} & 64.3\% (14) & 63.9\% (16) & \textbf{64.6\% (16)} & 62.8\% (17) \\
		\multicolumn{1}{c}{}    &              
		\multicolumn{1}{l}{Arrhythmia} & 81.3\% (39) & 81.8\% (44) & \textbf{81.9\% (46)} & 80.3\% (45) \\
		\multicolumn{1}{c}{}    &
		\multicolumn{1}{l}{EEG} & 90.1\% (37) & 90.7\% (27) & \textbf{91.2\% (33)} & 87.7\% (32) \\		
		\cmidrule{1-6}
	\end{tabular*}
	}
	\hspace{11mm}	
\end{table*}

\begin{table*}[!htb]
	\centering
	\setlength\tabcolsep{1.5pt} % default value: 6pt
	\caption{AUC-ROC $\&$ F1-Score Corresponding to the Peak Accuracy of SURI and GSA using Random Forest}
			\label{T16}
	\label{table_3}
	\resizebox{5in}{!}{
	\begin{tabular*}{\linewidth}{c c c c c c c c c c c}
		\cmidrule{1-11}
		&& \multicolumn{3}{c}{\head{AUC-ROC}} & \multicolumn{6}{c}{\head{F1-Score}}\\ \cmidrule{3-6}  \cmidrule{8-11} 
		%                & & \multicolumn{5}{c}{AUC-ROC} {F1-Score} \\ \cmidrule{3-8} \cmidrule{10-15} 
		&& GSA & SURI($\beta$=0.1) & SURI($\beta$=0.2) & SURI($\beta$=0.3)& &GSA & SURI($\beta$=0.1) & SURI($\beta$=0.2) & SURI($\beta$=0.3) \\ \cmidrule{3-6} \cmidrule{8-11}
		\multicolumn{1}{c}{\multirow{5}{*}{\begin{sideways}\parbox{2cm}{\centering Data Set \\ Name}\end{sideways}}}   &
		\multicolumn{1}{l}{Alizadeh}& 0.806 & 0.808 & \textbf{0.813} & 0.810 & & 0.901 & 0.904 & \textbf{0.909} & 0.901  \\
		\multicolumn{1}{c}{}    &
		\multicolumn{1}{l}{Breast Cancer} &0.945 &\textbf{0.946} & 0.944 & 0.937 & &0.938 & 0.935 & \textbf{0.942} &0.936  \\
		\multicolumn{1}{c}{}    &
		\multicolumn{1}{l}{SPECTF Heart} &\textbf{0.612} &0.608 &0.611 &0.603 & &0.723 & \textbf{0.724} & 0.718 &0.703 \\
		\multicolumn{1}{c}{}    & 
		\multicolumn{1}{l}{Arrhythmia} & 0.765 & 0.763 & 0.767 & \textbf{0.771}& &0.742 & 0.756 & \textbf{0.763} &0.749\\
		\multicolumn{1}{c}{}    &
		\multicolumn{1}{l}{EEG} &0.897 &0.901 &\textbf{0.905} &0.887 & &0.902 & 0.905 & \textbf{0.909} &0.893 \\		
		\cmidrule{1-11}
	\end{tabular*}
	}
	\hspace{11mm}
\end{table*}
\clearpage

\section{Additional Data}
\label{sec:AppendixB}
\begin{table}[h]
\centering
\caption{Feature Description for UCI Heart Disease Data Set}
\begin{tabular}{| c | l |}\hline
\label{TDiscri}
%\cmidrule{1-2}
\head{Feature Number} & \ \head{Feature Description} \\ \hline
%\cmidrule{1-2}
Feature 0 & \ Age in years (age) \\ \mygrayline
Feature 1 & \ Sex (sex)   \\ \mygrayline   
Feature 2 & \ Chest pain type (cp)       \\ \mygrayline
Feature 3 & \ Resting blood pressure on admission to the hospital (trestbps)\\  \mygrayline
Feature 4 & \ Serum cholestoral (chol)     \\ \mygrayline
Feature 5 & \ Fasting blood sugar (fbs)       \\ \mygrayline
Feature 6 & \ Resting ECG results (restecg) \\  \mygrayline
Feature 7 & \ Maximum heart rate during Thalium stress test (thalach) \\ \mygrayline
Feature 8 & \ Exercise induced angina (exang)     \\ \mygrayline
Feature 9 & \ ST depression induced by exercise relative to rest (oldpeak)  \\ \mygrayline
Feature 10 & \ Slope of the peak exercise ST segment (slope)     \\ \mygrayline
Feature 11 & \ Number of major vessels colored by flouroscopy  (ca)       \\ \mygrayline
Feature 12 & \ Thalium stress test result (thal)      \\ \arrayrulecolor{black}\hline
%\cmidrule{1-2}
\end{tabular}
\end{table}

\begin{figure}[h]
	\begin{center}
		\centerline{\includegraphics[width= 0.8\textwidth]{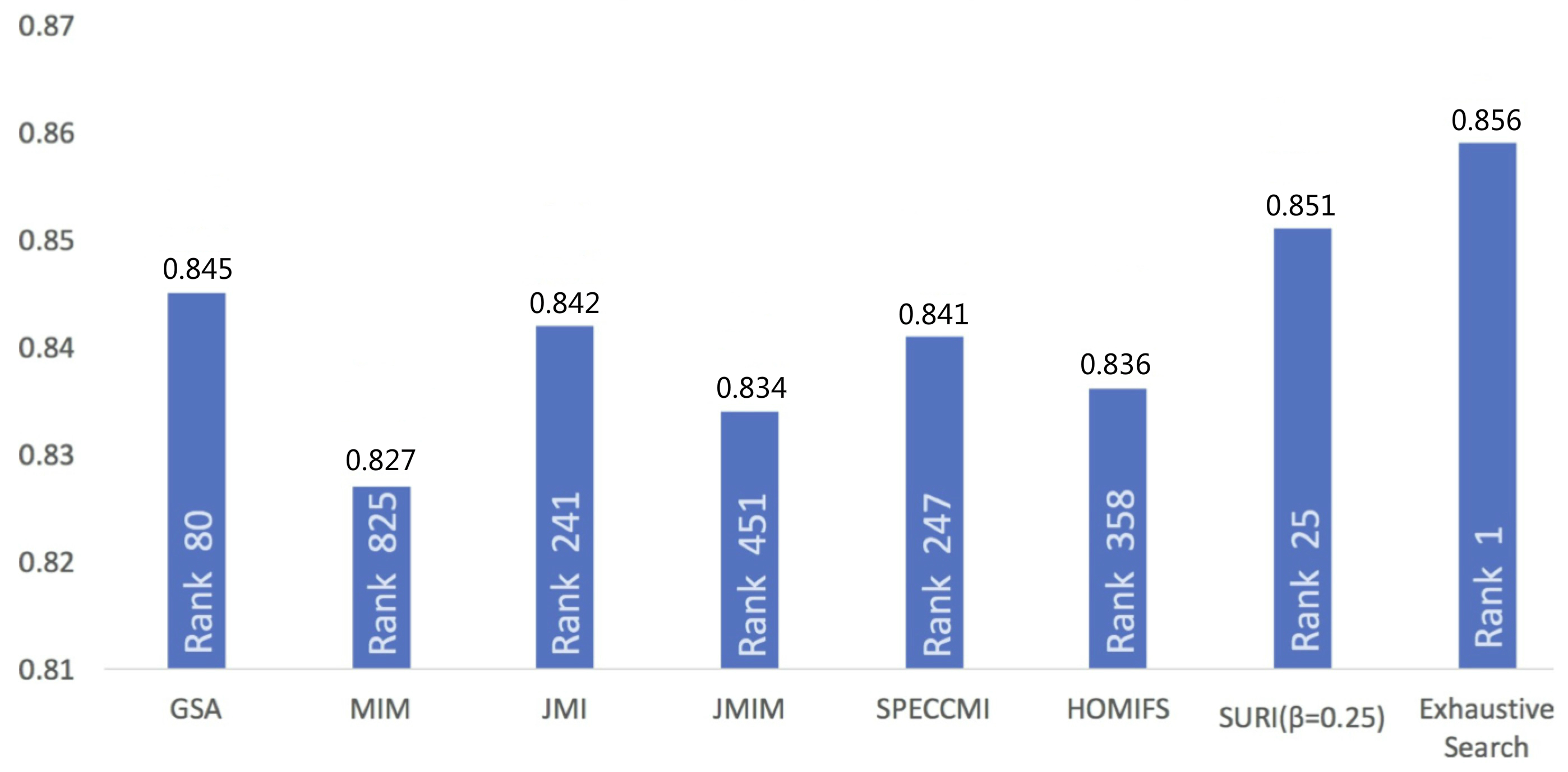}}
		\caption{SVM average peak accuracies of various MIBFS methods and corresponding ranks based on average peak accuracy.}
		\label{fig:bar}
	\end{center}\vspace{-7mm}
	%\vskip -0.3in
\end{figure}

\clearpage

\end{document}